\documentclass[pmlr]{jmlr}
\usepackage{longtable}
\usepackage{multirow}
\usepackage{booktabs}
\usepackage[load-configurations=version-1]{siunitx} 
\usepackage{amssymb}

\makeatletter
\def\set@curr@file#1{\def\@curr@file{#1}} 
\makeatother
\definecolor{mygray}{gray}{.9}

\theorembodyfont{\upshape}
\theoremheaderfont{\scshape}
\theorempostheader{:}
\theoremsep{\newline}

\jmlrvolume{149}
\jmlryear{2021}
\jmlrworkshop{Machine Learning for Healthcare}

\title[Multi-Label Generalized Zero Shot Learning for Chest Radiographs]{Multi-Label Generalized Zero Shot Learning for the Classification of Disease in Chest Radiographs}

\author{\Name{Nasir Hayat}
      \Email{nh2218@nyu.edu}\\ 
      \addr Engineering Division\\
      NYU Abu Dhabi, Abu Dhabi, UAE
      \AND
      \Name{Hazem Lashen}
      \Email{hl3372@nyu.edu}\\ 
      \addr Engineering Division\\
      NYU Abu Dhabi, Abu Dhabi, UAE
      \AND
      \Name{Farah E. Shamout}
      \Email{fs999@nyu.edu}\\ 
      \addr Engineering Division\\
      NYU Abu Dhabi, Abu Dhabi, UAE} 

\begin{document}

\maketitle

\begin{abstract}
Despite the success of deep neural networks in chest X-ray (CXR) diagnosis, supervised learning only allows the prediction of disease classes that were seen during training. At inference, these networks cannot predict an unseen disease class. Incorporating a new class requires the collection of labeled data, which is not a trivial task, especially for less frequently-occurring diseases. As a result, it becomes inconceivable to build a model that can diagnose all possible disease classes. Here, we propose a multi-label generalized zero shot learning (CXR-ML-GZSL) network that can simultaneously predict multiple seen and unseen diseases in CXR images. Given an input image, CXR-ML-GZSL learns a visual representation guided by the input's corresponding semantics extracted from a rich medical text corpus. Towards this ambitious goal, we propose to map both visual and semantic modalities to a latent feature space using a novel learning objective. The objective ensures that (i) the most relevant labels for the query image are ranked higher than irrelevant labels, (ii) the network learns a visual representation that is aligned with its semantics in the latent feature space, and (iii) the mapped semantics preserve their original inter-class representation. The network is end-to-end trainable and requires no independent pre-training for the offline feature extractor. Experiments on the NIH Chest X-ray dataset show that our network outperforms two strong baselines in terms of recall, precision, f1 score, and area under the receiver operating characteristic curve. Our code is publicly available at: \url{https://github.com/nyuad-cai/CXR-ML-GZSL.git}

\end{abstract}

\section{Introduction} Deep learning has enabled the development of computer aided diagnostic systems that can classify diseases in medical images with human level precision \citep{Qin2018ComputeraidedDI, chexnext, chestemerg}. One limitation of these networks is that they require a large amount of annotated data for training, which is often a tedious and expensive process that requires expert knowledge. Along with this, it becomes even more challenging to collect a significant amount of data for naturally occurring rare diseases or novel outbreaks, such as COVID-19 \citep{covidzsl}. Therefore, it becomes impractical to annotate a substantial amount of training data for all possible diseases in order to train a deep learning network. Due to the data limitations, at inference, a deep learning network is unable to classify disease classes that were unseen during training \citep{zsl}. However, radiologists can recognize previously unseen diseases by relying on the disease attributes learned through medical literature. 

\emph{Zero shot learning} (ZSL) has the potential to mimic the radiologist's behavior by recognizing unseen diseases using other sources of information \citep{zsl}. It is arguably the most extreme case of learning with minimal supervision. In brief, given a query image, ZSL methods find the correspondence between the visual representation of the image and its semantic representation \citep{xianCVPR17}. ZSL has achieved impressive results on natural images \citep{deeptag}. Nevertheless, most of the proposed methods assign only a single label to the each query image, and in many implementations, the assigned label can only be drawn from the seen classes only \citep{zsl,Long_2017_CVPR}. In medical imaging tasks, such as the classification of disease in chest X-ray images, an image may contain more than one disease and the labels could be from either the seen or unseen classes \citep{wang2017chestxray}. This limits the direct application of existing single-label ZSL methods for medical multi-label classification tasks. 
On the other hand, \emph{Multi-label ZSL} (ML-ZSL) allows the assignment of multiple labels to each image. Under Multi-label \emph{Generalized} ZSL (ML-GZSL), the goal is to assign multiple labels to a query image that could be in both seen and unseen classes \citep{gzsl}.  Existing ML-GZSL approaches developed for natural images perform the nearest neighbor search in the semantic space:
\vspace{-2mm}
\begin{equation}
  P^c_x = \langle f_x, \mathcal{W} \rangle,
  \label{eqn:scores}
\end{equation}
where $\langle ,\rangle$ is the cosine similarity function, $f_x$ are the visual features of a query image $x$, and $\mathcal{W}$ are the semantic embeddings for all possible classes. Equation \ref{eqn:scores} computes the similarity scores between the visual features and the semantic embeddings of each possible class. The similarity score indicates the relevance of the label to the image. More appropriately, these approaches perform the nearest neighbor search by mapping the aggregated visual features to the semantic space \citep{deeptag,Huynh_2020_CVPR}. However, these methods extract a fixed visual representation of the image from a pre-trained visual encoder or a detection network. Additionally, projecting these extracted visual features to the semantic space shrinks the diversity of the visual information, which gives rise to inherent limitations such as the hubness problem \citep{hubness}. 

To overcome the aforementioned challenges, we propose the CXR-ML-GZSL network for the classification of disease in chest X-rays (commonly referred to as CXR). In summary, our network consists of a visual feature encoder and two projection modules for each modality, which are the visual features and semantic embeddings. These modules project the independent modalities to the common latent space where the correspondence among them can be established. We evaluate our proposed method on the NIH chest X-ray dataset \citep{wang2017chestxray} and the results demonstrate that CXR-ML-GZSL outperforms baseline models.  In this work, we present the following three contributions:

\begin{itemize}
   
  \item From a technical perspective, we design an end-to-end trainable network that jointly learns the visual representation and aligns it with the semantic representation. Our design does not require any offline training for the visual feature encoder. 
  \item  Additionally, we propose a novel learning objective for the CXR-ML-GZSL to optimize the learned representations. The objective function simultaneously learns the predictive distribution, ensures that the visual representations are well centered around the class semantics in the latent embedding space, and enforces a constraint to preserve the original representation of class semantics in the latent space.
  \item From a medical imaging perspective, to the best of our knowledge, we are the first to propose a ML-GZSL framework for the classification of disease in chest X-rays.
\end{itemize}

We summarize related work in Section \ref{sec:related-work}, the methodology in Section \ref{sec:methodology}, the experiments in Section \ref{sec:experiments}, the results in Section \ref{sec:results}, the discussion in Section \ref{sec:discussion}, and finally, the conclusion in Section \ref{sec:conclusion}. 

\section*{Generalizable Insights about ML in the Context of Healthcare}
Conventional multi-label chest X-ray classification models are constrained by the availability of data and its annotations. Our work overcomes the challenge of collecting large-scale annotated datasets by leveraging the use of rich medical literature, since it is the main knowledge source for all the discovered diseases by the medical community. This highlights the role of multi-modal learning in healthcare applications. Although we focus on chest X-rays, the network design can be potentially generalized to any medical imaging task, since the semantic encoder is task-agnostic. Improving the diagnosis of unseen diseases at inference has the potential to save patient lives.


\section{Related work}
\label{sec:related-work}
\subsection{Inductive \& transductive ZSL} ZSL classifies previously unseen classes during training by transferring knowledge from seen classes. It relies on class semantics to bridge the gap between seen and unseen classes. The semantics are either obtained through manually annotated class attributes and textual descriptions embedded in a high-dimensional space \citep{att_zsl,ZhangXG16a}, or by extracting the label's semantic vectors using Word2vec or Glove \citep{mikolov2013distributed,Elhoseiny}. Previously proposed ZSL training methods can be categorized into  \emph{inductive ZSL}, which strictly trains on the data related to seen classes only \citep{xianCVPR17}, and \emph{transductive ZSL}, which assumes that unlabeled visual examples for unseen classes are also available during training \citep{transductive}. However, transductive ZSL violates the assumption that unseen classes are not seen during training, as it is impractical to assume that visual data of unseen classes is available during training. Therefore, addressing the ZSL under the inductive setting would be more realistic and renders a practical solution for the classification of chest X-rays.

\subsection{Multi-label generalized zero shot learning}
Few of the existing work address multi-label classification with natural images. \cite{Lee_2018_CVPR} constructed structured knowledge graphs by exploiting the semantic relationships in WordNet \citep{wordnet} to assign labels for unseen classes. A recent work, deep0tag \citep{deeptag}, extracts a bag of localized patches using a pre-trained detection network, aggregates these proposals, and then maps them in the semantic space to find the correspondence between seen and unseen classes. This approach however, requires a large-scale dataset with bounding box annotations to pre-train FasterRCNN \citep{fasterrcnn, zsd} for meaningful proposal generation. \cite{Huynh_2020_CVPR} proposed a shared multi-attention model to learn class agnostic multi-attention features by extracting a set of cropped regions' features using a pre-trained CNN. These attention features are projected to the semantic space to find the relevance within labels. Although most of these proposed approaches perform reasonably well on natural images, they are either not directly applicable to chest X-ray classification or rely on offline visual feature extractors trained on ImageNet \citep{ILSVRC15}. As shown in the work by \cite{raghu2019transfusion}, transferring knowledge from ImageNet provides little to no gain for radiograph diagnosis. As such, it is equally important to learn a meaningful visual representation for the visual encoder of CXR-ML-GZSL. Here, we propose to jointly learn the visual representation and align these representation with the semantics in an end-to-end network.

\subsection{Deep learning for chest radiographs} Multi-label classification assumes that multiple labels could appear in a single image. This is quite common for the classification of chest X-rays, where each image could contain multiple diseases. Previously, \cite{rajpurkar2017chexnet} trained a Densenet-121 \citep{densenet} by formulating a standard multi-label classification problem. However, their approach does not capture meaningful correlations among the different classes. \cite{Guan2018DiagnoseLA} used an attention mechanism by cropping the regions of interest extracted from attention heat maps provided by a global image encoder, in order to train a local network. They eventually fuse the features after the last pooling layers from both the global and local branches to assign classification scores. \cite{yao2018learning} used a DenseNet to encode a global representation of the X-rays and adopted a long-short term memory network to learn inter-dependencies among classes for better diagnosis \citep{lstm}. Despite their promising performance, these methods rely on large amounts of labeled data. Additionally, at inference, they can only predict classes that were seen during training.

There is very limited work on using ZSL for chest radiography. Recently, \cite{MVSE} proposed a GZSL framework using a two-branch autoencoder that combines external knowledge from three textual resources: chest X-ray reports, CT reports, and manually defined visual traits. Their design has a few limitations. First, the network predicts a single label, although the datasets have multiple labels per image. Second, it is transductive and hence assumes that the reports and unlabeled images for unseen cases are available during training, which contradicts the unseen assumption. Third, their semantic representation is learned from radiology and CT reports of already observed cases and is guided by manually annotated class attributes. Therefore, due to this data dependency on reports and manual annotations, its application is confined towards a closed set of seen and already observed unlabeled unseen classes. Unlike the work by \cite{MVSE}, our network can assign multiple labels to a query image and can generalize to an open set of unseen classes since it does not rely on existing reports and manual annotations.

\begin{figure}[ht!]
    \centering
    \includegraphics[width=.99\textwidth]{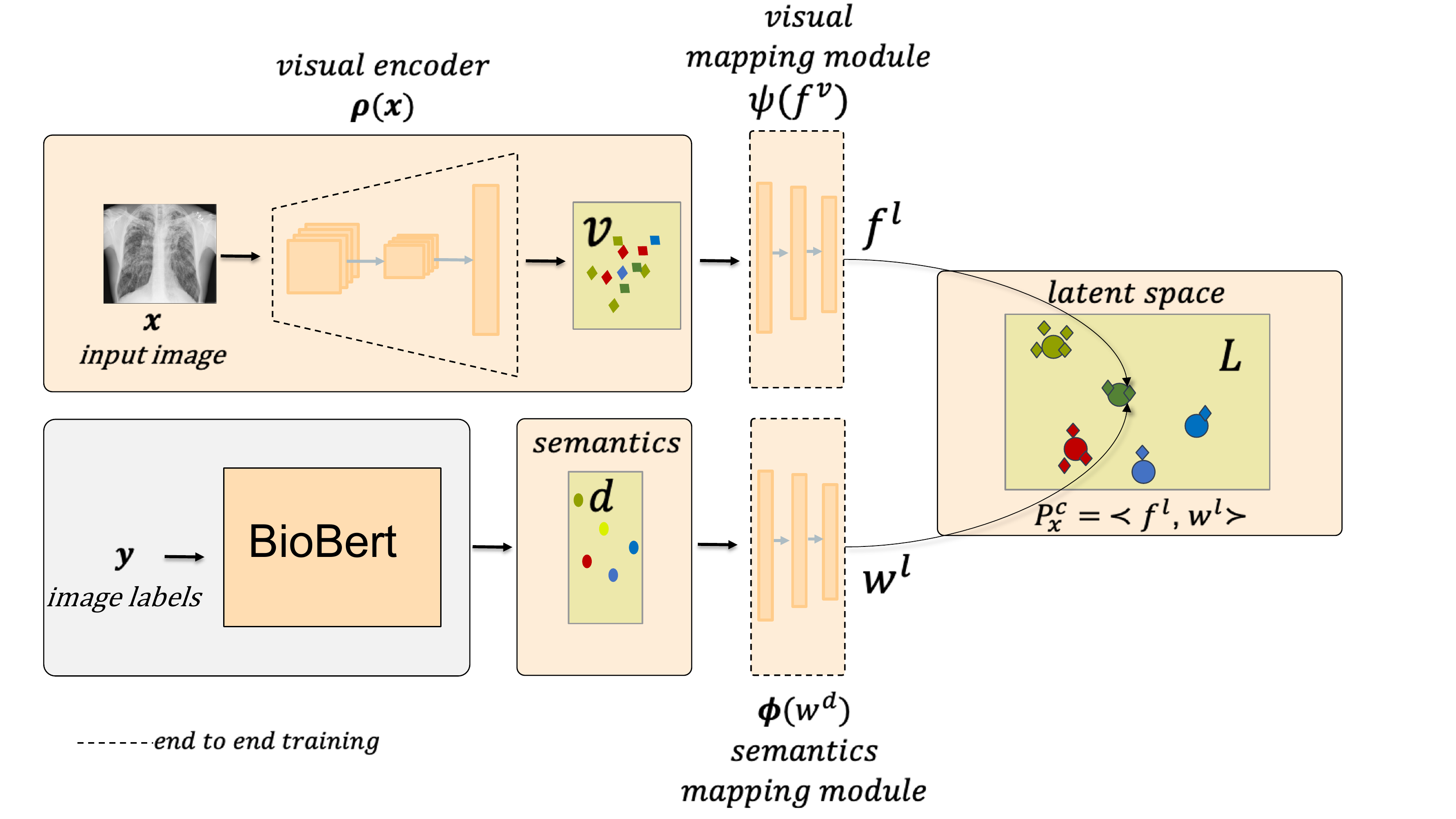}
    \vspace{-1mm}
    \caption{\footnotesize{\textbf{Overview of our CXR-ML-GZSL model for learning visual representations of chest X-rays.} An overview of our network for chest X-ray images. It includes a trainable visual encoder and $v$- and $d$-dimensional visual and semantics spaces, respectively. For an input image $x$ and its labels $y$, the network learns a visual representation guided by semantics extracted by BioBert. We perform end-to-end training of the visual encoder, visual mapping module, and the semantics mapping module, as indicated by the black dashed line. }}
    \label{fig:main_fig}
    \vspace{-3mm}
\end{figure}

\section{Methodology}
\label{sec:methodology}
\subsection{Problem formulation} Consider a set $ \mathcal{X}^{s}$ that contains the training images of seen classes only. Each $x \in \mathcal{X}^{s}$ is associated with a set of labels $ \textbf{y}_x$, where $ y_x^i \in \mathcal\{0,1\}_{i=1}^S$ and one indicates the presence of the $i^{th}$ disease among the $S$ seen classes during training. We denote the corresponding seen and unseen classes as $\mathcal{Y}^{s} = \{1,\dots, S\}$ and $\mathcal{Y}^{u} = \{S+1,\dots, C\}$, respectively, where $C$ is the total number of seen and unseen classes. Note here that both of these sets are disjoint such that $\mathcal{Y}^{s} \cap  \mathcal{Y}^{u} = \emptyset $. The training objective is to learn the visual representation of $x$ guided by the semantics of its labels $ \textbf{y}_x$. At inference, given a test image $x_{test}$, the goal is to predict $\textbf{y}_{x_{test}}$, where $y_{x_{test}}^i\in [ 0,1 ]_{i=1}^{C}$. In the following sections, we describe the proposed network's architecture and the learning objective.


\subsection{Network architecture}
An overview of the network architecture is visualized in Figure~\ref{fig:main_fig}. It consists of a trainable visual encoder, fixed semantic encoder, and alignment modules. We provide more details about each component and the loss function in this section.

\noindent \textbf{Visual encoder:}  To learn the visual representation, we first define a visual encoder $\boldsymbol{\rho}(x): \mathbb{R}^{w \times h \times c}  \mapsto\! \mathbb{R}^{v}$ which computes $f^v$, a $v$-dimensional visual representation of the input image $x$. Subsequently, a sub-module, referred to as the \textit{visual mapping module}, processes the visual representation $\boldsymbol{\psi}(f^v): \mathbb{R}^{v}  \mapsto\! \mathbb{R}^{l}$ and maps it to a latent manifold $l$, which is jointly learned with the semantic information.

\noindent \textbf{Semantics:} Let the semantic embeddings for seen classes be $\mathcal{W}^{s} = \{w_1, w_2, \cdots ,w_S\}$, where $w_i$ is a $d$-dimensional semantic representation for a class label $i \in \mathcal{Y}^{s}$, and it is extracted from the penultimate layer of BioBert \citep{Lee_2019} for all the trainable labels. To the best of our knowledge, our work is the first to adopt BioBert for ZSL problems in healthcare. Our preference to use BioBert was because of the fact that it learns effective contextualized word embeddings specific to the biomedical corpora. 

We define, $\boldsymbol{\phi}(w^d): \mathbb{R}^{d}  \mapsto\! \mathbb{R}^{l}$, referred to as the \textit{semantics mapping module}, which learns to map a $d$-dimensional semantic embedding to the joint latent manifold $l$. 

\noindent With respect to the proposed architecture, we redefine Equation \ref{eqn:scores} as: 
\begin{equation}
  P_x^S = \langle \boldsymbol{\psi} ( \boldsymbol{\rho} (x)),\   \boldsymbol{\phi}(\mathcal{W}^{s}) \rangle, 
  \label{eqn:scoresours}
\end{equation}
\noindent where $P_x^S$ represents the relevance scores of the training labels, i.e. $ P_x^S = \{p_1, p_2, \cdots ,p_S\}$. Those relevance scores represent the similarity between the visual input and each possible label.

\noindent \textbf{Training objective:} We formulate a training objective to jointly optimize the parameters of the network as follows:
\begin{equation}
    \min_{\boldsymbol{\phi} ,\boldsymbol{\rho} ,\boldsymbol{\psi}} \mathcal{L} = \mathcal{L}_{{rank}} +\gamma_{1} \mathcal{L}_{align} +\gamma_{2} \mathcal{L}_{con},
    \label{eqn:full_loss}
\end{equation}
where $\gamma_1$ and $\gamma_2$ are the regularization parameters for the loss components $\mathcal{L}_{align}$ and $\mathcal{L}_{con}$, respectively. In the next sections, we formally define each of the loss components. 

\subsection{\texorpdfstring{$\mathcal{L}_{rank}$}: Ranking loss of relevance scores} 

During training, the network assigns the relevance scores $P_x^S = \{p_1, p_2, \cdots ,p_S\}$ for each of the $S$ seen classes for an input image $x$. Given the ground-truth labels $ \textbf{y}_x$, where $y_x^i \in \mathcal\{0,1\}_{i=1}^S$, we denote $Y_p$
as the set of positive ground-truth labels (i.e., diseases that are present in the image $y_x^i=1$) and $Y_n$ as the set of negative ground-truth labels (i.e., diseases that are not present in the image $y_x^i=0$). Let us assume that $p_{y_p}$ and $\ p_{y_n}$ are the computed scores for a positive label and a negative label, respectively. Since this is a multi-label classification task, ideally, we wish to meet two conditions. First, $p_{y_p}$ should be higher than $p_{y_n}$, and second, the difference between $p_{y_p}$ and $p_{y_n}$ should be at least a margin value of $\delta$. 
Therefore, we formulate an image-level penalty score using the margin-based ranking loss as follows:   
\begin{equation}
    \mathcal{L}(P_x^S, \textbf{y}_x) = \frac{1}{S}\sum_{y_p \in Y_p } \sum_{y_n \in {Y_n}} max(\delta + (p_{y_n} - p_{y_p}) ,\ 0).
    \label{eq:rankingloss}
\end{equation}
In the above formulation, the loss is 0 if $p_{y_p}>p_{y_n}$ by a margin of at least $\delta$. It would inflict a penalty of at least $\delta$ when $p_{y_p}\leq p_{y_n}$. For all training samples, we average the image-level penalty scores over all images:


\begin{equation}
    \mathcal{L}_{{rank}} = \frac{1}{N} \sum_{\forall x \in \mathcal{X}^{s}} \mathcal{L} (P_x^S, \textbf{y}_x),
\end{equation}
where $N$ is the total number of images.

\subsection{\texorpdfstring{$\mathcal{L}_{align}$}: Alignment loss of visual-semantic representations}
To align the visual representations with their relevant semantic representations during training, we formulate an alignment loss for the two modalities as follows:
\begin{equation}
    \mathcal{L}_{align} = \frac{1}{N} (1 - \sum_{\forall x \in X^s} \langle \boldsymbol{\psi} (\boldsymbol{\rho} (x)),\  \boldsymbol{\phi}(w_x) \rangle),
\end{equation}
where $w_x$ is the corresponding semantic embeddings for the input image $x$ and $\langle , \rangle$ defines the cosine similarity function. In case multiple labels are present in the image, then their semantic embeddings are averaged in $ w_x $. The averaging allows the alignment of the visual representation with its semantics in the multi-label setting. 


\subsection{\texorpdfstring{$\mathcal{L}_{con}$}: Semantics inter-class consistency regularizer}
The semantic and visual representations stem from two separate modalities. The semantic representations are learned through medical textual literature, while the visual representations are learned from chest radiograph. To bridge the gap between both modalities, we learn two mapping functions that project the representations to a common domain. While the visual representations are fine-tuned during training, the semantic representations are fixed after being extracted from the natural language encoder. Mapping the semantic representations to the latent space may lead to a loss in inter-class relevance. Hence, we aim to ensure that the semantic manifold remains consistent throughout the projection process by leveraging the semantic inter-class relationship. To achieve this, we design an $L_1$ regularization term within the learning objective which considers the similarity between the classes between the original space and the projected space, such as: 
\begin{equation}
        \mathcal{L}_{con} = \sum_{w_i \in W } \sum_{\substack{w_j \in W \\ j\neq i}} \| \langle {w_i},\   {w_j} \rangle - \langle {\boldsymbol{\phi}(w_i)},\  {\boldsymbol{\phi}(w_j)} \rangle \| 
    \end{equation}
where $w_i$ and $w_j$ are the original semantic representations for two classes and $\boldsymbol{\phi}(.)$ computes their projected representations. Ideally, the cosine similarity between the two classes in the original space should be the same in the projected space, and hence the loss would be 0. 


\subsection{Inference} After training the network, we obtain the fine-tuned image encoder $\boldsymbol{\rho}(x)$ and mapping modules $\boldsymbol\psi(f^v)$ and $\boldsymbol\phi(w^d)$, which project both the encoded visual features and the semantic embeddings to the joint latent space, respectively. At inference, for a test image $x \in \mathcal{X}^{C}$, we update $\mathcal{W}$ in Equation \ref{eqn:scoresours} to include both seen and unseen semantic embeddings. Contrary to conventional multi-label classification, after a simple modification of $\mathcal{W}$, we can get the prediction scores $ P_x^c = \{p_1, p_2, \cdots ,p_c\}$ for a set of $C$ classes comprising both seen and unseen, such that:    
\begin{equation}
  P^C_x = \langle \boldsymbol\psi (\boldsymbol\rho (x)),\  \boldsymbol\phi (\mathcal{W}^{C}) \rangle.
  \label{eqn:infer}
\end{equation}
In Equation \ref{eqn:infer}, $\mathcal{W}^{C}$ represents the updated $\mathcal{W}$ and includes the semantic embeddings of both seen and unseen classes. 
\section{Experimental settings}
\label{sec:experiments}

\subsection{Dataset}
To evaluate the proposed network CXR-ML-GZSL, we ran the experiments on the publicly available NIH Chest X-ray dataset  \citep{wang2017chestxray}. The dataset contains 112,120 frontal X-ray images from across 30,805 unique patients. We split the dataset randomly to obtain a training set ($70\%$), validation set ($10\%$), and test set ($20\%$). Each image is associated with 14 possible classes. We include all 14 classes and perform a random split of seen (10 in total) and unseen (4 in total) classes. The seen classes were Atelectasis, Effusion, Infiltration, Mass, Nodule, Pneumothorax, Consolidation, Cardiomegaly, Pleural Thickening, and Hernia, and the unseen classes were Edema, Pneumonia, Emphysema, and Fibrosis. We strictly exclude all images from the training set that are associated with any unseen class label, in accordance with the inductive setting. The final training set contained 30,758 images, the validation set contained 4,474 images, and the test set contained 10,510 images.

\subsection{Model training and selection}
In this section, we describe the experimental setting for CXR-ML-GZSL. To encode the visual information, our method is designed to work with any state-of-the-art image convolutional neural network. We ran all of our experiments with Densenet-121, since it has shown the best results for chest X-ray classification \citep{rajpurkar2017chexnet}. To do so, we removed the final classification layer and used the resulting network as the visual encoder $\boldsymbol\rho(x)$ that yields the visual representation $f^v \in \mathbb{R}^{1024}$. 

The visual mapping module was parameterized as a three layer feed-forward neural network formally defined as
$\boldsymbol{\psi}: f^v \xrightarrow{} \texttt{fc1} \xrightarrow{} \texttt{Relu}  \xrightarrow{} \texttt{fc2} \xrightarrow{} \texttt{Relu} \xrightarrow{} \texttt{fc3} \xrightarrow{} {f^l}$, where $\texttt{fc1}$ is a fully connected layer which learns the weight matrix  $\mathbf{W}_{\texttt{fc1}} \in \mathbb{R}^{1024 \times 512}$ and a bias vector $\mathbf{b}_{\texttt{fc1}} \in \mathbb{R}^{512}$. The subsequent layer learns the weight matrix  $\mathbf{W}_{\texttt{fc2}} \in \mathbb{R}^{512 \times 256}$ and a bias vector  $\mathbf{b}_{\texttt{fc2}} \in \mathbb{R}^{256}$. The last layer learns the weight matrix  $\mathbf{W}_{\texttt{fc3}} \in \mathbb{R}^{256 \times 128}$ and a bias vector $\mathbf{b}_{\texttt{fc3}} \in \mathbb{R}^{128}$, which is eventually projected into the joint latent space with the semantic embeddings. The semantic mapping module follows a symmetric architecture to visual mapping module. Our design of the mapping modules enables the network to work with embeddings extracted from different architectures, which are eventually mapped to a common space irrespective of the original dimensions.

We trained the network using the Adam optimizer \citep{kingma2017adam} for a total of 100 epochs while reducing the initial learning rate ($lr$) by a factor of 0.01 once the validation loss stagnates for 10 epochs. It took about 8 hours to train a single model on an NVIDIA Quadro RTX 6000 GPU. Through out our experiments, we set $\delta=0.5$ in Equation \ref{eq:rankingloss}. To optimize the other hyperparameters ($\gamma_1$, $\gamma_2$ in Equation \ref{eqn:full_loss}, and the learning rate), we ran multiple experiments by randomly selecting $\gamma \in \{0.1, 0.01, 0.05\}$ and $lr \in \{ 0.0001, 0.00005, 0.00001\}$ and then chose the best performing model on the validation set in terms of the mean harmonic AUROC. The grid search ranges were selected based on the results of preliminary experiments. Our codebase was developed using the Pytorch deep learning library \citep{pytorch}.


\subsection{Performance metrics}
We used evaluation metrics that are commonly reported in ML-GZSL methods \citep{Lee_2018_CVPR, Huynh_2020_CVPR}. We computed overall precision, recall and f1 scores for the top $k$ predictions where $k \in \{2, 3 \}$ for GZSL. We selected a smaller value of $k$ since our dataset has fewer number of classes compared to natural imaging datasets \citep{nus_wide_civr09, openimages}. Additionally, we report the average area under the receiving operating characteristic curve (AUROC) for seen and unseen classes and its harmonic mean, since computing recall for top $k$ predictions may not be a sufficient indicator of class-wise performance. It is important to note that the harmonic mean measures the inherent bias of GZSL methods towards the seen classes.

\begin{table}[ht!]
 \caption{\footnotesize{\textbf{Performance evaluation of the network compared to state-of-the-art baselines on the test set.} Multi-label GZSL performance on the NIH chest X-ray dataset in terms of recall (\textbf{r}), precision (\textbf {p}) and \textbf {f1} scores at the top k predictions. We report the \textbf{AUROC} for seen (\textbf{S}) classes, unseen (\textbf{U}) classes, and their harmonic mean (\textbf{H}). Our proposed end-to-end trained network CXR-ML-ZSL is denoted as ``$OUR_{e2e}$". The numbers in \textbf{bold}  are the best values in the respective column. 
 } } \vspace{-3mm}
    \centering
    \resizebox{.7\textwidth}{!}{
    \begin{tabular}{c c c c| c c c | c c c }
        \toprule & 
         \multicolumn{3}{c}{\bfseries k=2}  & \multicolumn{3}{c}{\bfseries k=3} & \multicolumn{3}{c}{\bfseries AUROC} \\
        \bfseries Method & \bfseries r@k & \bfseries p@k & \bfseries f1@k & \bfseries r@k & \bfseries p@k & \bfseries f1@k & \bfseries S & \bfseries U & \bfseries H \\
        \midrule
        $LESA$ (\citeyear{Huynh_2020_CVPR}) & 0.14 & 0.10 & 0.03 & 0.21 & 0.11 & 0.05 & 0.51 & 0.50 & 0.50\\
        $MLZSL$ (\citeyear{Lee_2018_CVPR}) & 0.20 & 0.19 & 0.16 & 0.30 & 0.17 & 0.20 & 0.72 & 0.54 & 0.62 \\ 
        \hline
        
        $OUR_{e2e}$& \textbf{0.36} & \textbf{0.33} & \textbf{0.32} & \textbf{0.47} & \textbf{0.28} & \textbf{0.34} & \textbf{0.79} & \textbf{0.66} & \textbf{0.72}\\
        \bottomrule
    \end{tabular}}
    \vspace{-3mm}
    \label{tab:recall}
\end{table}

\begin{table}[ht!]
\caption{\small \footnotesize{\textbf{Class-wise AUROC comparison across all disease classes in the test set.}} In this table, we show the performance AUROC across all seen and unseen classes. Our proposed method ($OUR_{e2e}$) shows significant gains for a number of individual classes. Compared with the second best method (MLZSL), our method shows an AUROC gain of $22.2\%$ for the mean performance for unseen classes. The \underline{\textit{italicized}} classes are unseen and the \textbf{bold} numbers represent the best AUROC in the respective column.} 
\centering \vspace{-3mm}
\resizebox{.99\textwidth}{!}{
    {
        \begin{tabular}{c|c|c|c|c|c|c|c|c|c|c|c|c||c|c|c|c}
        \toprule[0.1em]
            {\textbf{{Method}}}&\rotatebox{90}{{seen mean}}&\rotatebox{90}{{unseen mean}}&\rotatebox{90}{{atelectasis}}&\rotatebox{90}{{cardiomegaly}}&\rotatebox{90}{{effusion}} & \rotatebox{90}{{infiltration}}&\rotatebox{90}{{mass}}&\rotatebox{90}{{nodule}}&\rotatebox{90}{{pneumothorax}}&\rotatebox{90}{{consolidation}}&\rotatebox{90}{{pleural }}\rotatebox{90}{{thickening}}&\rotatebox{90}{{hernia}}&
            \rotatebox{90}{\underline{\textit{pneumonia}}}&\rotatebox{90}{\underline{\textit{edema}}}&\rotatebox{90}{\underline{\textit{emphysema}}}&\rotatebox{90}{\underline{\textit{fibrosis}}}\\
            \toprule[0.1em]
            $LESA$ (\citeyear{Huynh_2020_CVPR}) & 0.51 & 0.52 & 0.54 & 0.51 & 0.48 & 0.50 & 0.52 & 0.50 & 0.51 & 0.54 & 0.52 & 0.47 & 0.50 & 0.54 & 0.50 & 0.53 \\
            $MLZSL$ (\citeyear{Lee_2018_CVPR}) & 0.72 & 0.54 & 0.70 & 0.76 & 0.76 & 0.64 & 0.67 & 0.69 & 0.76 & 0.68 & 0.66 & \textbf{0.90} & 0.52 & 0.51 & 0.58 & 0.54\\
        
            \midrule
            $OUR_{e2e}$& \textbf{0.79} & \textbf{0.66} & \textbf{0.76} & \textbf{0.90} & \textbf{0.83} & \textbf{0.70} & \textbf{0.80} & \textbf{0.75} & \textbf{0.83} & \textbf{0.69} & \textbf{0.72} & \textbf{0.90} & \textbf{0.62} & \textbf{0.67} & \textbf{0.74} & \textbf{0.60}\\
            \bottomrule
        \end{tabular}}
    }
    \vspace{-3mm}

\label{tab:class_wise}
\end{table}

\section{Results}
\label{sec:results}
\subsection{Comparison to baseline models}
We compared the performance of proposed approach ($OUR_{e2e}$), with two state-of-the art ML-GZSL methods: LESA \citep{Huynh_2020_CVPR} and MLZSL \citep{Lee_2018_CVPR}. Table \ref{tab:recall} summarizes the performance results on the test set. The results show that our method performs best across all metrics, reaching an AUROC of 0.66 on the unseen classes and a harmonic AUROC of 0.72 across both seen and unseen classes. LESA achieved the worst performance, while MLZSL achieved the second best performance. In particular, our method outperforms MLZSL by a significant margin, with a percentage change of $73.68 \%$ in terms of precision@2, for example.

Table \ref{tab:class_wise} compares the AUROC values for each class against the state-of-the-art methods. Our method achieves the best performance across all seen classes compared to the baseline methods, except for Hernia, where it achieves a comparable performance to MLZSL (0.90 AUROC). Across the unseen classes, it achieves the best AUROC performance compared to both baseline methods.  

\begin{figure}[t]
    \centering
   
    \includegraphics[width=.325\textwidth]{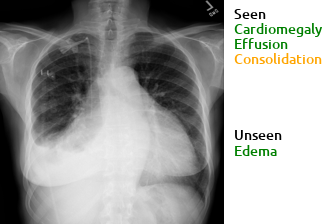}
    \includegraphics[width=.325\textwidth]{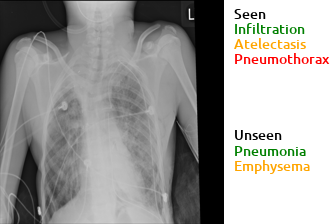}
    \includegraphics[width=.325\textwidth]{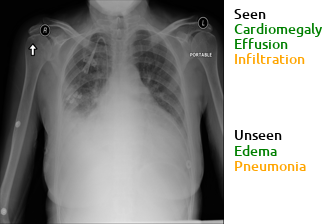} \\
    \vspace{0.2cm}
    \includegraphics[width=.325\textwidth]{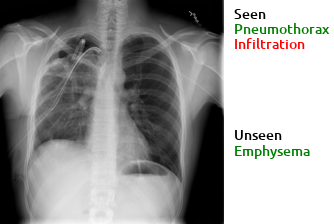}
    \includegraphics[width=.325\textwidth]{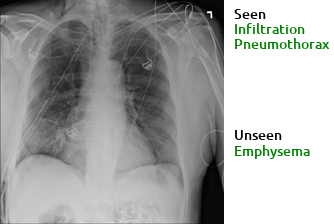} 
    \includegraphics[width=.325\textwidth]{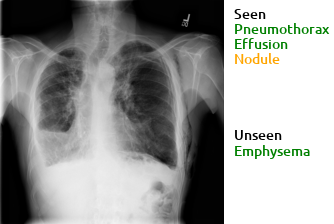} \\
    \vspace{0.2cm}
    \includegraphics[width=.325\textwidth]{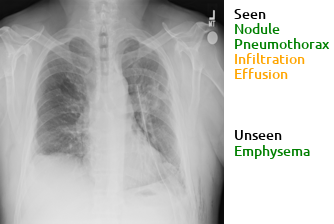}
    \includegraphics[width=.325\textwidth]{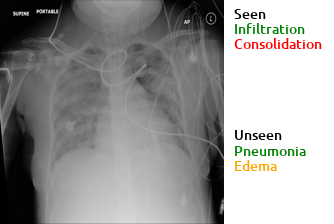}
    \includegraphics[width=.325\textwidth]{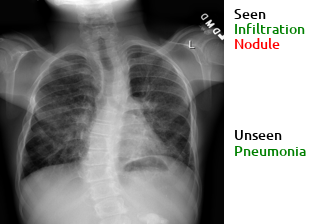}
    
    \caption{\footnotesize{\textbf{Qualitative results for the top three predictions per image within the test set.} The labels in green, orange and red represent true positives, false negatives, and false positives respectively. The green and orange labels combined represent the ground-truth labels.}}
    \label{fig:qual}
\end{figure}

Figure \ref{fig:qual} shows example predictions for 9 test chest X-ray images. We select the top three predictions on each test image. We observe that our method is still able to predict the unseen classes even when the number of ground-truth labels is high, such as greater than 3. This demonstrates the effectiveness of the proposed method under challenging settings of simultaneously predicting multiple seen and unseen classes.


   
    

\subsection{Ablation studies} We performed two ablation studies using the validation set. In all ablation studies, we set the initial learning rate to 0.0001, $\gamma_{1}=0.01$, and $\gamma_{2}=0.01$. Table \ref{tab:ablation} shows the AUROC for seen and unseen classes and their harmonic mean with different formulations of the objective function. In particular, we ran experiments to evaluate the individual contribution of incorporating the alignment loss and the semantic consistency regularizer. While the performance of seen classes remains consistent through out the experiments (0.783-0.791 AUROC), an improvement in the unseen AUROC can be seen when incorporating the alignment loss $\mathcal{L}_{{align}}$, which ensures that the learned visual representations are centered around the class semantics. Additionally, preserving a consistent semantic manifold using the $\mathcal{L}_{{con}}$ regularizer contributed significantly to improving the performance among the unseen classes. Therefore, we used the loss component that incorporated all of the ranking loss, alignment loss, and the semantics regularizer to train the final model.

We also investigated the impact of using an offline visual encoder compared to our proposed end-to-end training approach. We ran experiments by freezing the visual encoder pre-trained on ImageNet for the purpose of extracting the visual features. Then, we separately trained the visual encoder on the NIH Chest X-ray dataset with seen classes only and eventually used it to extract fixed visual features. Table \ref{tab:training_approaches} shows the performance of these approaches compared to our end-to-end training approach. It is interesting to note that the end-to-end approach performs reasonably well compared to the other approaches. This emphasizes the point that learning a well-aligned visual representation guided by the semantic embeddings helps in improving the performance, especially for unseen classes where the AUROC is boosted significantly.


\begin{table}[t]
\caption{\footnotesize{\textbf{Ablation study for the loss function using the validation set.}} The individual contribution of each loss component towards the performance in terms of AUROC of multi-label generalized ZSL. We progressively integrate each component, improving the performance across unseen classes, while the performance of seen classes remains persistent. The \textbf{bold} numbers represent the best AUROC in the respective column.}
    \centering
    \resizebox{.6\textwidth}{!}{
    \begin{tabular}{*6{c}}
    \toprule
    \bfseries $\mathcal{L}_{{rank}}$ & \bfseries $\mathcal{L}_{{align}}$ & \bfseries $\mathcal{L}_{{con}}$ &  \bfseries Seen &  \bfseries Unseen &  \bfseries Harmonic Mean \\
    \midrule
    \checkmark
 & &  & 0.790 & 0.574 & 0.665\\
    \checkmark
 & \checkmark
 &  & \textbf{0.791} & 0.601 & 0.683 \\
    \checkmark
 & \checkmark
 & \checkmark
 & 0.783 & \textbf{0.663} & \textbf{0.718} \\
    \bottomrule
    \end{tabular}}
    \label{tab:ablation}
\end{table}

\begin{table}[t]
 \caption{\footnotesize{\textbf{Comparison of training strategies on the validation set}}. We compare different training strategies and report the AUROC for our proposed method compared to offline visual encoders trained on ImageNet and the NIH chest X-ray datasets. End-to-end training achieves the best performance across the seen and unseen classes. The \textbf{bold} numbers represent the best AUROC in the respective column.} 
    \centering \resizebox{.6\textwidth}{!}{
    \begin{tabular}{c c c c }
        \toprule
        \bfseries Training method & \bfseries Seen &  \bfseries Unseen &  \bfseries Harmonic Mean \\
        \midrule
       Pre-training with Imagenet &  0.721 & 0.608 & 0.660 \\ 
        Pre-training with NIH & 0.771 & 0.609 & 0.681 \\
        \midrule
        End-to-end training  & \textbf{0.783} & \textbf{0.663} & \textbf{0.718} \\
        \bottomrule
    \end{tabular}}
    \label{tab:training_approaches}
\end{table}

\section{Discussion}
\label{sec:discussion}
Recent advancements in the field of deep learning for medical imaging are highly dependent upon the availability of large-scale datasets. In this study, we present a promising approach to develop a multi-label diagnostic network that can classify unseen classes using generalized zero shot learning. Our CXR-ML-GZSL network leverages the contextualized semantics learned through rich medical literature and learns visual representations guided by these semantics, with a unique learning objective. We train our model for a set of seen classes and then evaluate it on a set of held-out seen and unseen classes using the NIH chest X-ray dataset. The assumption is that the unseen classes are never exposed to the model during training, in order to mimic a real-life scenario of classifying unseen ``rare" diseases. The results show that the proposed network generalizes well to both seen and unseen classes and achieves impressive performance gains over previous state-of-the-art methods. Note that for clinical deployment, we suggest that the clinical practitioner is shown the list of diseases ranked from most-to-least likely based on the prediction scores. Clinicians can also define a threshold to convert the prediction scores ($P_x^c$) to binary using sensitivity and specificity analysis.


A recent work by \cite{MVSE} proposed and evaluated a GZSL method for chest X-ray imaging. They include a subset of 9 classes out of 14 based on the availability of CT reports. Out of these, they randomly select 6 classes as seen and the remaining 3 are used as unseen classes. However, their assumption of the availability of unseen data during training (including CT reports, radiology reports \& chest radiographs) violates the unseen characteristic for ZSL and may not give meaningful insight on the performance for unseen classes. Therefore, to fully assess the robustness of GZSL towards unseen classes, we strictly ensure that no auxiliary data related to unseen classes is used during the training. Due to evident differences in assumptions on the training settings, we are unable to directly compare the results between our work and the work of \cite{MVSE}.

\subsection*{Limitations} To evaluate the proposed network, we are restricted to using an existing dataset. This has several limitations. First, the dataset has a fewer number of classes compared to natural imaging benchmarks. To evaluate the robustness of our proposed method and future works, we emphasize on the need for creating a more challenging benchmark dataset with a sufficiently large number of classes. In addition, we randomly selected seen and unseen classes to evaluate the method for chest X-ray classification. Although this evaluation is based on the assumption that the four classes were unseen, the method needs to be further evaluated with other seen and unseen splits, potentially on other medical tasks using different types of imaging, using data from other institutions to assess for generalizability, and on standard computer vision benchmarks. Additionally, we perform hyperparameter tuning for the learning rate and $\gamma$ parameters only. We expect the results to further improve if we tune other hyperparameters, like batch size and dimensions of mapping modules.





\section{Conclusion}
In this work, we propose a multi-label generalized zero shot learning (CXR-ML-GZSL) network for chest X-ray classification. Through our experimentation, we show that at inference, our network can simultaneously assign multiple labels both from seen and unseen classes. Since our network is strictly restricted to be trained on seen classes and no auxiliary information in the form of either unlabeled chest X-rays or clinical reports related to unseen classes are provided during training, we believe that CXR-ML-GZSL has substantial potential to diagnose unseen classes at inference especially for rare or emerging diseases that are challenged by the lack of annotated data. 
\label{sec:conclusion}


\bibliography{sample.bib}

\end{document}